\pdfoutput=1

\documentclass[11pt]{article}

\usepackage[]{acl} 

\usepackage{times}
\usepackage{latexsym}
\usepackage{makecell}
\usepackage{multirow}
\usepackage{colortbl}
\usepackage{graphicx}
\usepackage{amsmath}
\usepackage{amsfonts}
\usepackage{algorithm}
\usepackage{algorithmic}

\usepackage{booktabs}
\usepackage{multirow}
\usepackage{multicol}

\usepackage[T1]{fontenc}

\usepackage[utf8]{inputenc}

\usepackage{microtype}

\title{	
 Knowledgeable Salient Span Mask for Enhancing \\Language Models as Knowledge Base}

\author{Cunxiang Wang\textsuperscript{$1$,$2$}, Fuli Luo\textsuperscript{$3$}, Yanyang Li\textsuperscript{$4$}, Runxin Xu\textsuperscript{$5$}, Fei Huang\textsuperscript{$3$} and Yue Zhang\textsuperscript{$1$\thanks{\ \ The corresponding author} }
\\
\textsuperscript{$1$}School of Engineering, Westlake University; \textsuperscript{$2$}Zhejiang University;
\\
\textsuperscript{$3$}Damo Academy, Alibaba Group;
\textsuperscript{$4$}CUHK;
\textsuperscript{$5$}Peking University\\
\textsuperscript{}\{wangcunxiang, zhangyue\}@westlake.edu.cn}

\begin{document}
\maketitle

\begin{abstract}
Pre-trained language models (PLMs) like BERT have made significant progress in various downstream NLP tasks. However, by asking models to do cloze-style tests, recent work finds that PLMs are short in acquiring knowledge from unstructured text. 
To understand the internal behaviour of PLMs in retrieving knowledge, we first define knowledge-baring (K-B) tokens and knowledge-free (K-F) tokens for unstructured text and ask professional annotators to label some samples manually. Then, we find that PLMs are more likely to give wrong predictions on K-B tokens and attend less attention to those tokens inside the self-attention module.Based on these observations, we develop two solutions to help the model learn more knowledge from unstructured text in a fully self-supervised manner.
Experiments on knowledge-intensive tasks show the effectiveness of the proposed methods. To our best knowledge, we are the first to explore fully self-supervised learning of knowledge in continual pre-training.

\end{abstract}



\section{Introduction}

Pre-trained language models (PLMs), such as BERT~\citep{BERT} and GPT~\citep{GPT}, have greatly improved the performance of many NLP tasks in the past few years. Pre-training has been regarded as a promising way for acquiring common knowledge from unstructured plain text. However, how to effectively learn knowledge for PLMs is still an unsolved problem \citep{LAMA, wang-etal-2021-generative}, especially in those tasks which need explicit usage of knowledge.
There are mainly two common ways to enhance PLMs with more knowledge. One is to introduce structured knowledge bases \citep{ERNIE-THU,K-ADAPTER} while the other uses unstructured text.
Compared with structured knowledge bases, unstructured text is easier to acquire and construct. In addition, with freer format, unstructured text can express better complex knowledge.
\begin{figure}[t]
  \center{
  \includegraphics
  [width=7cm]  
  {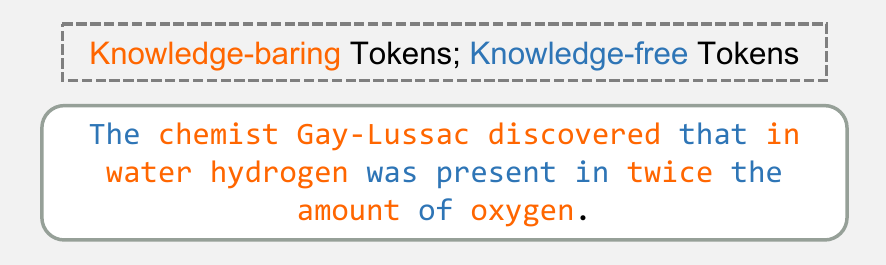}}
  \caption{Examples of knowledge-baring (K-B) tokens and knowledge-free (K-F) tokens.}
  \label{example}
  \vspace{-2mm}
\end{figure}

\begin{figure}[t]
  \center{
  \includegraphics
  [width=7cm]  
  {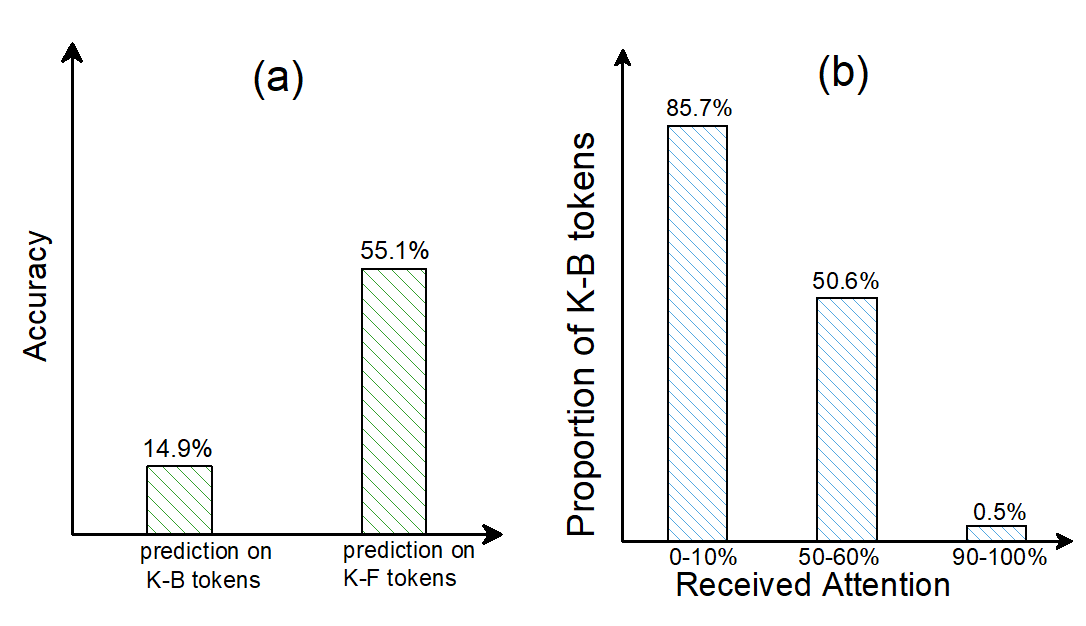}}
  \caption{The RoBERTa's behavior  on probing samples: (a) the model performs worse on knowledge-baring tokens than on knowledge-free tokens; (b)  knowledge-baring tokens are likely to receive less attention in the self-attention process.}
  \label{two_findings_figure}
  \vspace{-3mm}
\end{figure}



We focus on enhancing the ability of PLMs in acquiring knowledge from unstructured text. To ensure that the knowledge learned from text is factual knowledge and is valid for downstream tasks, we start with studying PLMs' behavior s on factual knowledge. 
First of all, we explore which tokens in the text embody factual knowledge in a more fine-grained manner (i.e., token-level). This not only helps us better understand the model's behavior  of memorizing and utilizing knowledge, but also motivates us to design methods for better acquiring knowledge. 
In particular, for a piece of text, the tokens which are essential for humans to understand the text's factual knowledge are considered as \textit{knowledge-baring}; otherwise, they are \textit{knowledge-free}. One example is presented in Figure~\ref{example}.
We analyze PLMs' behavior s on knowledge by manually annotating whether each token in samples is knowledge-baring.
As shown in Figure~\ref{two_findings_figure} (a), we find that PLMs perform worse on knowledge-baring tokens in the cloze-style test.
In addition, shown in Figure~\ref{two_findings_figure} (b), the transformer-based model is likely to gain less attention on knowledge-baring tokens.

Intuitively, for better acquiring knowledge from unstructured text, the model is expected to mask-recover more knowledge-baring words when trained on the unstructured text and receive less influence from knowledge-free words.
To this end, based on our observation in Figure~\ref{two_findings_figure} (Section~\ref{sec:knowledge_section}), we propose two solutions at the mask policy and attention levels of the PLM:
(1) We have two mask policies. The first is to perform random masking on the training corpus before each training iteration and find out which masks the model fails to predict correctly. These incorrectly predicted tokens are regarded as knowledge-baring tokens for masking in this training iteration. 
The second is to use the attention to determine which tokens are more likely to be knowledge-baring for masking.
(2) At the attention level, we adopt the visibility matrix to prevent knowledge-free tokens from affecting other tokens during self-attention. 

Extensive experiments are conducted on three tasks. Specifically, to check whether the model has learned the knowledge from unstructured text, we let the model perform on the LAMA Probing task \citep{LAMA}, a standard cloze-style test. To test whether the model can utilize the learned knowledge, we also introduce two probing task, namely Closed-book QA \citep{SQuAD} and Knowledge Graph Reasoning \citep{TREx}.
Note that there is no labelled data for finetuning for the three tasks, they are only used for probing how much knowledge the model has learned from unstructured text. Besides, the training corpus contains all needed knowledge of evaluation and testing. The test examples of three tasks are presented in Table~\ref{tasks_table} in the Appendix.
Experiments on the three tasks show the effectiveness of the proposed methods, achieving up to 6.1 and 5.5 points absolute improvement in the LAMA Probing task on two datasets, up to 6.7 points absolute improvement in the Closed-book QA task and 2.6 points absolute improvement in the KG Reasoning task. 
%

To our knowledge, we are the first to explore the relationship between PLMs' behavior  and knowledge in the token-level and the first to research fully self-supervised learning of knowledge in continual pre-training.

\section{Background}

\subsection{Backbone Model}
\label{sec:backbone}
We choose the RoBERTa \citep{RoBERTa} model as our backbone model, and the original pre-training objective of RoBERTa as our baseline.
The RoBERTa model is built on the encoder of the Transformer model\citep{transformer}. For each layer of RoBERTa, it consists of a multi-head self-attention layer and a position-wise feed-forward network.
For $i_{th}$ layer, the self-attention output of $j_{th}$ head is
\begin{equation}
    A_j = \mathrm{softmax}(\frac{Q_jK_j^{T}}{\sqrt{d_k}})V_j
    \label{A_j}
\end{equation}
where $Q, K, V$ are the query, key and value vectors, respectively. $d_k$ is the dimension of $Q, K, V$ vectors.

\subsection{Pre-training and Downstream Tasks}
\label{sec:pretraining_and_tasks}
To a PLM with parameters $\theta$, if it is continually pre-trained corpus $\mathbb{C}$ with masked tokens $V^{*}$. Its optimized parameters should be
\begin{equation}
    \Tilde{\theta} = \mathop{\arg\max}\limits_{\theta} (P(V_p = V^{*}| \mathbb{C}; \theta))
\end{equation}
where $V_p$ is the predicted tokens by $\theta$ on the masked corpus $\mathbb{C}$.

For a task with input and label $(x_i, y_i) \in  (X, Y^*)$. The ideal parameters of the PLM should be 
\begin{equation}
    \Bar{\theta} = \mathop{\arg\max}\limits_{\theta} ( P(Y_p = Y^{*} | X; \theta))
\end{equation}
where $Y_p$ is the predictions by $\theta$ on $X$.

In this work, for the downstream task, we only consider the data whose needed knowledge appears in the pretraining corpus $\mathbb{C}$. Thus, we make an approximation that the distribution of $X$ can be seen as a special subset of corpus $\mathbb{C}$, so we have
\begin{equation}
    P(X) \in P(\mathbb{C})
    \label{subseteq}
\end{equation}

To achieve high performance on downstream tasks, we match pre-training-optimized $\Tilde{\theta}$ to task-ideal $\Bar{\theta}$. Since we can only influence $V^{*}$ when the dataset $\mathbb{C}, X, Y^*$ are determined, we can expect $V^{*}$ to be similar to $Y^{*}$.

Due to the diversity of downstream task types and needed knowledge, there is only one commonality we can confirm that all $(x_i, y^*_i) \in (X, Y^*)$ have in common, that is they all contain at least one piece of factual knowledge.

However, it is difficult to identify all factual knowledge from $\mathbb{C}$ with existing NLP tools\footnote{For example, an NER tagger. Named entities are only a small set of factual knowledge.}. So, we expect PLMs to identify factual knowledge by self-supervised training. Thus, we try to learn the behavior of how PLMs recognizing knowledge.

\section{Behavior of PLMs in Knowledge} \label{sec:knowledge_section}

To better probe how PLMs learn knowledge from unstructured text, we start to identify the type and role of each word.
Inspired by knowledge graphs as well as our observations, we find that knowledge in a sentence is largely embodied by a few keywords. For the remaining words, even if they are deleted, we can still receive the factual knowledge the sentence conveys. 
\begin{itemize}
    \item \textbf{knowledge-baring}: For a given text, if the deletion of one token will make it relatively hard for humans to obtain the factual knowledge contained in the text correctly, we take the token as knowledge-baring;
    \item \textbf{knowledge-free}: For a given text, if the deletion of one token still allows humans relatively easy to obtain the factual knowledge contained in the text correctly, we take the token as knowledge-free.
\end{itemize}

One example is shown in Figure~\ref{example}. Note that knowledge-free tokens are not totally free of knowledge. They certainly have some kind of knowledge, such as linguistic and semantic knowledge. They are just relatively less important to the factual knowledge, which we emphasize in this work.

We randomly sample 100 cases from the LAMA SQuAD dataset and LAMA Google RE dataset \citep{LAMA}, respectively and then use the tokenizer of RoBERTa to tokenize each sentence. We ask three annotators, who are all Ph.D. students, manually label each token as \textbf{knowledge-baring} and \textbf{knowledge-free}. The inter-annotator agreement for samples of LAMA SQuAD/LAMA Google RE is 0.920/0.938, respectively.
The statistic of labelled tokens is shown in Table~\ref{token_number}.

We also use the Stanford CoreNLP toolkit \citep{CoreNLP} to conduct part-of-speech tagging analysis on those samples. We find that the most knowledge-baring tokens are nouns (64.2\%), verbs (11.6\%), numbers (9.2\%) and adjective words (6.5\%)  while most knowledge-free tokens are preposition or subordinating conjunctions (25.1\%), comma and punctuation (23.6\%), determiners (15.2\%) and verbs (11.7\%) for the two sets of samples. We also put the detailed results in the Appendix Table~\ref{pos_on_samples}. From the results, we can see that we do not limit the scope of knowledge to entities or nouns. We expand it to nouns, verbs, numbers, adjective words, etc.

To better understand the model's behavior  on comprehending knowledge, we mainly explore two questions: (1) Does the model perform better on knowledge-baring contents or knowledge-free contents? (2) Can the model's attention scores reveal its association with knowledge?

\begin{table}[t]
  \centering
  \small
  \setlength{\tabcolsep}{0.52mm}
  \begin{tabular}{c|c|c}
  \toprule
  & \makecell{ number of \\K-B Tokens} &\makecell{number of \\K-F tokens} \\
  \midrule
  \makecell{LAMA SQuAD} & 739 & 532\\
  \midrule
  \makecell{LAMA Google RE} & 1715
  & 975\\
  \bottomrule
  \end{tabular}
  \caption{The number of tokens that are knowledge-baring and knowledge-free we have labelled for the samples of the two dataset.
  }
  \label{token_number}
  \vspace{-3mm}
\end{table}

\subsection{Accuracy on Knowledge-Baring and Knowledge-Free Tokens}
\label{sec:accuracy_analysis}
To investigate the first question, we first mask each token of the sentences in both datasets. For example, if one sentence contains 10 separate tokens, we derive 10 sentences with ``<mask>'' on each token after processing this sentence. If one word is tokenized to several tokens, we mask those tokens together. The detail is shown in the Table~\ref{tab:analysis_data} (a) in Appendix.
Then, we ask the model to predict the mask(s) of processed sentences.

To better understand the influence of pre-training on model learning knowledge, we use the original PLM as well as the continued pre-trained model to predict on the processed sentences.  For continual pre-training, we first find the Wikipedia snippets where the sentences are from and then train the model using the pre-training objective with the snippets for 100 iterations. 

The performances of RoBERTa and continued pre-trained RoBERTa on two types of tokens on two datasets are presented in Table~\ref{acc_on_each}.
From the result, we find that the  model performs much worse on knowledge-baring tokens than on knowledge-free tokens, which is 14.9\% to 55.1\% on SQuAD and 38.6\% to 83.4\% on Google RE. Even if the model is continual pre-trained, the accuracy of knowledge-baring tokens is still lower than that of K-F tokens, which is 39.2\% to 82.8\% on SQuAD and 67.2\% to 93.5\% on Google RE.
The results show that it is more difficult for models to learn factual knowledge from unstructured text than non-knowledge.

\begin{table}[t]
  \centering
  \small
  \setlength{\tabcolsep}{0.52mm}
  \begin{tabular}{c|c|c}
  \toprule
   & \makecell{Knowledge-Baring} &\makecell{Knowledge-Free} \\
  \midrule
  RoBERTa-Orig & 
  \cellcolor{red!15}14.9\%&\cellcolor{red!55}55.1\%\\
  \midrule
  \makecell{RoBERTa-Cont} &
  \cellcolor{red!39}39.2\%&\cellcolor{red!83}82.8\%\\
  \midrule
  \multicolumn{3}{c}{(a) On the LAMA SQuAD Samples.}\\
  
  \midrule
   & \makecell{Knowledge-Baring} &\makecell{Knowledge-Frees} \\
  \midrule
  RoBERTa-Orig & 
  \cellcolor{red!27}38.6\%&\cellcolor{red!73}83.4\%\\
  \midrule
  \makecell{RoBERTa-Cont} &
  \cellcolor{red!57}67.2\%&\cellcolor{red!84}93.5\%\\
  \midrule
  \multicolumn{3}{c}{(b) On the LAMA Google RE samples.}\\
  \bottomrule
  \end{tabular}
  \caption{The probing accuracy on two types of tokens for original model (RoBERTa-Orig) and continued pre-trained model (RoBERTa-Cont) along with the original pre-training mask policy.
  Both models perform worse on knowledge-baring tokens.
  }
  \label{acc_on_each}
  \vspace{-3mm}
\end{table}

\begin{table*}[t]
  \centering
  \small
  \setlength{\tabcolsep}{0.6mm}
  \begin{tabular}{c|c|c|c|c|c|c|c|c|c|c|c}
  \toprule
  \makecell{}& 0$\sim$10\% & 10$\sim$20\% & 20$\sim$30\% & 30$\sim$40\%& 40$\sim$50\% & 50$\sim$60\%& 60$\sim$70\% & 70$\sim$80\%& 80$\sim$90\%& 90$\sim$100\% & Corr* \\
  \midrule
  Original RoBERTa
  &\cellcolor{red!86}85.7\%
  &\cellcolor{red!79}78.9\%
  &\cellcolor{red!72}72.3\%
  &\cellcolor{red!69}69.3\%
  &\cellcolor{red!58}58.1\%
  &\cellcolor{red!50}50.6\%
  &\cellcolor{red!46}46.4\%
  &\cellcolor{red!22}22.4\%
  &\cellcolor{red!6}5.5\%
  &\cellcolor{red!1}0.5\%
  & -1.0\\
  \midrule
  \makecell{RoBERTa-Cont} &\cellcolor{red!75}75.1\%
  &\cellcolor{red!73}72.8\%
  &\cellcolor{red!65}64.9\%
  &\cellcolor{red!65}65.0\%
  &\cellcolor{red!57}57.4\%
  &\cellcolor{red!53}53.3\%
  &\cellcolor{red!54}53.9\%
  &\cellcolor{red!41}40.7\%
  &\cellcolor{red!9}9.4\%
  &\cellcolor{red!1}0.5\%
  & -0.98 \\
  \midrule
  \multicolumn{12}{c}{(a) On LAMA SQuAD samples} \\
  \midrule
  & 0$\sim$10\% & 10$\sim$20\% & 20$\sim$30\% & 30$\sim$40\%& 40$\sim$50\% & 50$\sim$60\%& 60$\sim$70\% & 70$\sim$80\%& 80$\sim$90\%& 90$\sim$100\% & Corr* \\
  \midrule
  Original RoBERTa
  &\cellcolor{red!97}97.6\%
  &\cellcolor{red!92}92.2\%
  &\cellcolor{red!85}84.7\%
  &\cellcolor{red!75}75.4\%
  &\cellcolor{red!71}70.9\%
  &\cellcolor{red!59}59.1\%
  &\cellcolor{red!54}53.2\%
  &\cellcolor{red!43}42.9\%
  &\cellcolor{red!11}10.6\%
  &\cellcolor{red!5}4.9\%
  & -1.0\\
  \midrule
  \makecell{RoBERTa-Cont} 
  &\cellcolor{red!82}81.7\%
  &\cellcolor{red!78}77.6\%
  &\cellcolor{red!77}77.3\%
  &\cellcolor{red!76}75.8\%
  &\cellcolor{red!69}68.9\%
  &\cellcolor{red!63}63.4\%
  &\cellcolor{red!62}61.9\%
  &\cellcolor{red!50}50.2\%
  &\cellcolor{red!38}38.1\%
  &\cellcolor{red!6}5.8\%
  & -1.0\\
  \midrule
  \multicolumn{12}{c}{(b) On LAMA Google RE samples}\\
  \bottomrule
  \end{tabular}
  \caption{The relationship between knowledge-baring proportion (in red) and the level of receiving attention (the first row). The head X-Y\% indicates those tokens rank in bottom X-Y\% on attention receiving, for example, 0-10\% means those tokens receive least attention. The cell with red color is the K-B proportion of those tokens. RoBERTa-Cont is the continued pre-trained RoBERTa. 
  The last column is the the Spearman's rank correlation coefficient between the level of receiving attention and K-B proportion.
  We can see that tokens receiving more attention are less likely to be K-B.
  }
  \label{results_none_mask}
  \vspace{-3mm}
\end{table*}

\subsection{Attention on Knowledge-Baring and Knowledge-Free Tokens}
\label{sec:attention_analysis}
For the second question, we feed-forward the model on the sentences without masking them. For each token, we calculate the sum of all tokens' received attention weights and sum up for all layers and heads. The received attention (RcAtt) weight of token $t$ in the model is 
\begin{equation}
    RcAtt_{t} = \sum_{i=1}^{L} \sum_{j=1}^{H}\sum_{k=1}^{N}att_{ijkt}
    \label{RcAtt}
\end{equation}
where $L$ is the layer number, $H$ is the head number and $N$ is the token number; $att_{ijkt}$ means in $layer_i$ $head_j$, the attention score $token_k$ to $token_t$.

We sort all the tokens by their RcAtt scores for each sentence and divided them into 10 percent segments. Next, we calculate the proportion of knowledge-baring tokens in each segment. Same as the previous question, we not only use the original PLM to predict,  but also test the continued pre-trained model.

The results are presented in Table~\ref{results_none_mask}. 
We can see that the attention scores strongly correlate with whether the tokens are knowledge-baring. The K-B tokens are more likely to receive less attention, while the K-F tokens are more likely to receive more attention.
When the model is continual pre-trained, this phenomenon still exists but at a slightly reduced level.


\paragraph{Conclusions.} Based on the above two probing experiments, we can conclude that: (1) PLMs perform worse on knowledge-baring words (i.e., with higher prediction error); (2) The knowledge-baring words are more likely to receive less attention than knowledge-free ones.

\section{Methods}

In this section, we propose two methods based on the conclusion of the above probing experiments, making PLMs learn more knowledge from unstructured text.




\subsection{Mask Policy} \label{sec:mask_policy}

Corresponding to Section~\ref{sec:pretraining_and_tasks}, this section is to determine $V^*$ given pre-training corpus $\mathbb{C}$.
Initially, RoBERTa randomly chooses tokens from the input text to mask. However, recent work \citep{wang-etal-2021-generative} shows that it is inefficient to memorize knowledge with random masks. 
Therefore, we aim to enable the model to focus on learning knowledge-baring content.  
We do not provide any label information to the model, the model needs to find the K-B tokens from the input text without any supervision.

From the Section~\ref{sec:knowledge_section}, we find that the K-B tokens are related to whether the model can accurately predict the token and attention weight the token receive. Hence we provide two corresponding selective mask policies for the model to find and mask the K-B tokens. Note that the two selective mask policies are mutually exclusive, so we compare their performance rather than combine them.

\textbf{RoBERTa-Sel-I}. Since the model performs much worse on knowledge-baring tokens than on knowledge-free tokens, we can use this feature to find out K-B tokens from unstructured text. Before each training iteration, we randomly mask some tokens of the training text and predict on the masks, and then we \underline{\textbf{Sel}}ect out tokens that are \underline{\textbf{I}}naccurately predicted and treat them as K-B tokens.
Besides finding K-B tokens, this policy also helps the model to avoid learning those tokens which it has already learned previously.

\textbf{RoBERTa-Sel-A}. As the knowledge-baring tokens are more likely to receive less attention, we can make use of the attention score each token receives. Before each training iteration, we let the model forward on the non-masked training text, and then we calculate each token's attention weights, which is the same as Eq~\ref{RcAtt}. Next, we \underline{\textbf{Sel}}ect out the tokens that get the least \underline{\textbf{A}}ttention and treat them as K-B tokens.

After finding knowledge-baring tokens, we first randomly mask them and then randomly choose to mask all remaining tokens. For example, we set the first-phase mask language modelling (MLM) probability as 15\%, and second-phase MLM probability as 10\%, if the text has 100 tokens and we find 20 K-B tokens using one of our methods, we first mask 100$\times$15\%=15 tokens from the K-B tokens and then mask 100$\times$10\%=10 tokens from the left  85 tokens. The two-phase masks will be combined for pre-training.

Salient Span Mask (SSM) \citep{REALM} uses a trained NER tagger and a regular expression to identify named entities and date from the raw corpus. These salient spans are selected and masked. We also conduct the SSM experiments on our dataset as a comparison. 
But note that the SSM policy requires external labelled data to train a NER tagger and prior knowledge to design the expression while our methods are free of any external information. And SSM can only detect named entities and expressions while our methods can detect all kinds of knowledge.

\begin{figure}[t]
  \center{
  \includegraphics
  [width=6cm]  
  {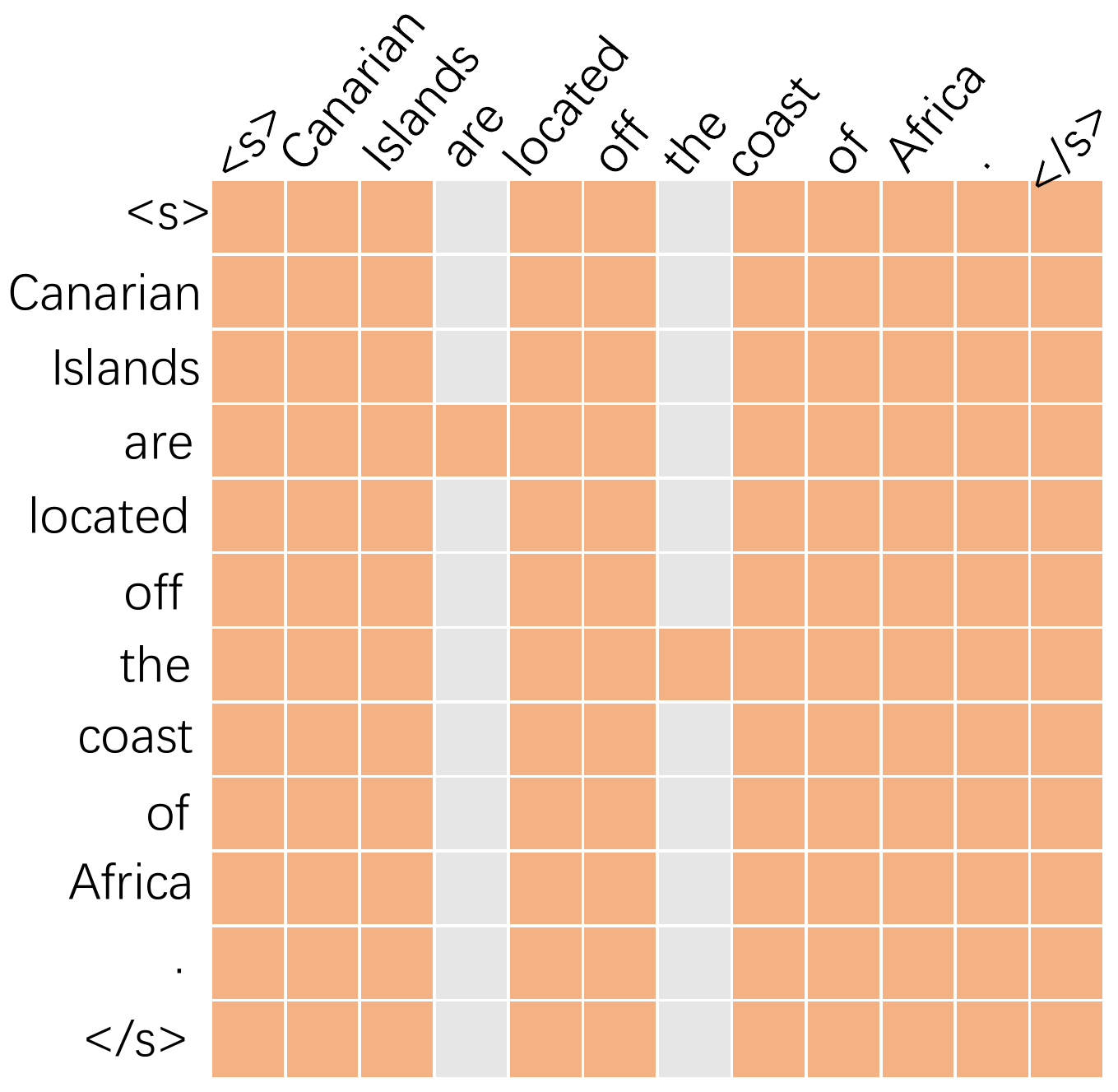}}
  \caption{Illustration of the visibility matrix. The orange square means the left token can see the top token while the gray square means it cannot. In this example, the token ``are'' and ``the'' are invisible to other tokens.}
  \label{vm_figure}
  \vspace{-3mm}
\end{figure}

\subsection{Visibility Matrix} \label{sec:visibility}

\begin{algorithm}[t]
\small
\renewcommand{\algorithmicrequire}{\textbf{Input:}}
\renewcommand{\algorithmicensure}{\textbf{Output:}}
    \begin{algorithmic}
    \STATE \textbf{Special Dataset Construction}: 
    \STATE (1) Forward RoBERTa on the training data.
    \STATE (2) Select tokens which receive the least 10\% attentions.
    \STATE (3) Ask the whole words which contain those tokens from the training corpus. 
    \STATE (4) The masked train set is served as the special dataset.
    \STATE \textbf{Initialization}: 
    \STATE (1) Set a positive real number threshold $\tau$.
    \STATE (2) Tokenize the special validation data, collect all tokenized tokens that appear more than $\tau$ times to a set $T$.
    \STATE (4) Add special tokens ``<s>'', ``</s>'', ``<pad>'' and ``<mask>'' to the set $T$.
    \STATE (5) Initialize the set $T_n$ as empty.
    \STATE (5) Evaluate the model accuracy $ACC_0$ on the special dataset.
    \FOR{token $t$ in set $T$}
    \STATE (1) Make $t$ invisible to other tokens
    \STATE (2) Evaluate the model accuracy $ACC$ on the special dataset.
    \IF{$ACC$ > $ACC_0$} 
    \STATE Add $t$ to $T_n$.
    \ENDIF
    \ENDFOR
    \RETURN $T_n$
    \end{algorithmic}
    \caption{Detecting ``harmful'' tokens.}
    \label{algorithm}
\end{algorithm}

In addition to making the model pay more attention to K-B tokens during the continual pre-training, we also consider making the model pay less attention to knowledge-free tokens.
To achieve this goal, we adopt the concept of visibility matrix from \citet{UniLM} and \citet{UniLMv2}. Using the visibility matrix, we expect those tokens that harm knowledge memorization cannot influence other tokens.


Figure ~\ref{vm_figure} is the illustration of the visibility matrix. 
During the self-attention process, if token $q$ can attend to token $p$, in other words, the hidden state of token $q$ can be influenced by the hidden state of token $p$, we consider token $q$ is visible to token $p$, otherwise, it is invisible. 
After adding visibility matrix mechanism to self-attention module, the self-attention output of $i$ layer and $j$ head in Eq~\ref{A_j} 
is changed to

\begin{equation}
    A_j = \mathrm{softmax}( \frac{Q_jK_j^{T}}{\sqrt{d_k}} + M^{*})V_j
\end{equation}
where $M^{*} \in \mathbb{R}^{n \times n}$,  $M^{*}_{pq}=-\infty$ if  token $q$ is visible to token $p$ and $M^{*}_{pq}=0$ if token $q$ is invisible to token $p$.

By conducting pilot experiments on making manually chosen irrelevant tokens invisible by other tokens, we find it effective to boost performance on the three tasks. 
So, we continue to design an algorithm to detect those tokens which will hurt the performance of the model. Since the training data does not have any label, we construct a special dataset from the training data to find the ``harmful'' tokens. 
The algorithm is presented in Algorithm~\ref{algorithm}. For each time, we make one token invisible, and check whether it will help the evaluation performance on the special dataset.

Note that there are three main differences between the proposed visibility matrix and the mask matrix used in recent works~\citep{UniLM,UniLMv2}:
1) The visibility matrix is independent on input masks while mask matrix only make the masked tokens invisible; 2) We have designed an automated algorithm to search invisible tokens rather than by random masking; 3) The invisible tokens can still see themselves while the tokens in mask matrix cannot.

For each task and dataset, we use Algorithm~\ref{algorithm} to find  ``harmful'' tokens automatically. In practice, we use the original RoBERTa-large model or the continued pre-trained RoBERTa-large model to evaluate. 
After finding those tokens, we make them invisible to all other tokens during training, validation and testing periods.
An example of the processed visibility matrix is shown in Figure~\ref{vm_figure}.


\section{Tasks}
We adopt three tasks in this work: LAMA probing, Closed-book QA, and Knowledge Graph (KG) Reasoning. 
The examples and detailed statistics of the three tasks are presented in Table~\ref{tasks_table} and Table~\ref{task_data} in the Appendix.
Note that inputs of three tasks are all in prompt-style.
And, for the three tasks, all needed knowledge of validation/test questions can be directly found in the training data.
These tasks are slightly different from ordinary tasks, as the training data is a corpus $\mathbb{C}$ and evaluation/test data is $X, Y^*$. This setting is to evaluate whether PLMs can learn knowledge from $\mathbb{C}$ and apply in downstream tasks. 
\footnote{Most knowledge-intensive tasks can be converted to such prompt-style. However, only in few of them we can map the needed knowledge of each case to a passage or a paragraph. That is why we use such tasks and datasets.}

The \textbf{LA}nguage \textbf{M}odel \textbf{A}nalysis Probing is a \citep{LAMA} task that \textit{directly} evaluate how much knowledge PLM obtain from unstructured text. For each example, the training case contains a passage and the validation/test case contains a cloze-style query and answer pair. The model needs to learn knowledge from training passages and use the knowledge to fill the ``<mask>'' tokens in the validation/test cloze-style sentences. 
The Closed-book QA task and the Knowledge Graph Reasoning task are to testify whether the PLM can utilize its learned knowledge in downstream tasks.

For each sample in the \textbf{Closed-book QA} task, the training case contains a sentence, while the validation/test case contains a cloze-style QA pair, whose question has one or several ``<mask>'' tokens after the ``$?$''. The model needs to fill the ``<mask>'' tokens. 
For each sample in the \textbf{KG Reasoning} task, the training case contains a sentence, while the validation/test case contains a cloze-style triple, whose object is replaced with one or several ``<mask>'' tokens. The model needs to fill the ``<mask>'' tokens. To make the model adapt to the cloze-style triples answer, for 20\% training sentences, we add the corresponding triple at the end of each sentence and remove the triple from the validation/test set.

\textbf{Data.} The task data originate from public released datasets, which are the SQuAD dataset and the Google RE dataset from the LAMA dataset. 
For the former, we link the probes to SQuAD1.1 dataset \citep{SQuAD} and find the related questions and passages of each case. Then we use the passages as training data and probes as the validation/test data to construct the dataset for LAMA Probing task. Moreover, we use the recovered probing sentences as the training data and the questions concatenated with ``<mask>'' as the validation/testing data for the Closed-book QA task.
For the latter, we use the snippet of each case as training data and probing sentences as the validation/test data. Furthermore, we use passages as the training data and use the <subject, relation, object> triples as the validation/test data for the KG Reasoning task.

\begin{table*}[t]
  \centering
  \small
  \begin{tabular}{c|cc|c|c|c}
  \toprule
   & LAMA SQuAD & LAMA Google RE & Closed-book QA & KG Reasoning & Average \\
  \midrule
  RoBERTa-Orig & 16.4 & 24.6 & 0.0 & 2.6 & 10.9 \\
  RoBERTa-Cont &  33.6 (+0.0) & 58.4 (+0.0)  &  37.9 (+0.0) & 28.1 (+0.0) & 39.5 (+0.0) \\
\midrule
  RoBERTa-SSM & 37.5 (+3.9)& 62.6 (+4.2) & 42.7 (+4.8)& \textbf{31.2 (+3.1)} & 43.5 (+4.0)\\
  \midrule
  RoBERTa-Sel-A & 35.9 (+2.3) & 62.4 (+4.0) & 44.4 (+6.5)& 27.7 (-0.4) & 42.6 (+3.1) \\

  RoBERTa-Sel-I & \textbf{39.7 (+6.1)}&63.5 (+5.1) & 43.6 (+5.7) & 29.5 (+1.4) & 44.1 (4.6)\\
  \midrule
  RoBERTa-Cont-VM & 38.5 (+4.9) & 62.8 (+4.4) & 43.4 (+5.5) & 29.6 (+1.7) & 43.6 (+4.1)\\
  RoBERTa-Sel-I-VM &37.2 (+3.6) &\textbf{63.9 (+5.5)} & \textbf{44.8 (+6.7)} & 30.7 (+2.6) & \textbf{44.2 (+4.8)}\\
    
  \bottomrule
  \end{tabular}
\caption{The accuracy on three knowledge intensive tasks. The first block denotes the results of original and continued pre-trained RoBERTa. The second and third blocks show the performance of improved models in terms of \underline{\textbf{Sel}}ective mask policy (Section~\ref{sec:mask_policy}) and \underline{\textbf{V}}isibility \underline{\textbf{M}}atrix (Section~\ref{sec:visibility}). The numbers in brackets show the absolute improvements compared to the continued pre-trained RoBERTa.}
\vspace{-2mm}
  \label{results}
\end{table*}

\begin{table}[t]
  \centering
  \small
  \setlength{\tabcolsep}{0.52mm}
  \begin{tabular}{c|c|c}
  \toprule
   & \makecell{LAMA-SQuAD} &\makecell{LAMA-Google RE} \\
  \midrule
  \makecell{RoBERTa-Orig} & 13.9\% & 38.6\%\\

  \makecell{RoBERTa-Cont} & 38.4\% &67.2\%\\
  \midrule
  \makecell{RoBERTa-Sel-A} & 41.8\% &71.4\%\\

  \makecell{RoBERTa-Sel-I} & 42.6\% &71.6\%\\
  
  \makecell{RoBERTa-Cont-VM} & 41.9\% &71.0\%\\
  \bottomrule
  \end{tabular}
  \caption{The probing results on the annotated knowledge-baring tokens.
  }
  \label{acc_on_kb}
  \vspace{-3mm}
\end{table}

\section{Experiments}

\paragraph{Settings.}
We adopt the RoBERTa-large model as our base model, and conduct continual pre-training on it.
We follow most of the traditional pre-training hyper-parameters of RoBERTa \citep{RoBERTa}, such as training batch size, optimization method and model configurations. However, some specific parameters are modified when applying our methods. We present needed hyper-parameters at Section~\ref{sec:hyper_para} in the Appendix.




\subsection{Overall Results}
Table~\ref{results} shows the results on three tasks. 
Our methods beat baseline models on average scores of four datasets, which gains up to 33.3 and 4.8 points of advantage over original RoBERTa-Large and RoBERTa-Large continually pre-trained with random masking, respectively. And all of our methods outperform baselines, showing effectiveness of them.

In detail, firstly, we investigate the masking policy (Section~\ref{sec:mask_policy}) in continual pre-training. It can be found that our proposed two selective mask policies (RoBERTa-Sel-I and RoBERTa-Sel-A) outperform the original random mask policy (RoBERTa-Cont), obtaining up to 6.1/5.1, 6.5 and 1.4 absolute improvement on three tasks, respectively. It indicates that our methods can enhance the RoBERTa with more domain specific knowledge in the continual pre-training process.

Furthermore, we find that model trained with Visibility Matrix (VM) mechanism (Section~\ref{sec:visibility}) can substantially achieve better accuracy. For example, RoBERTa-Cont-VM outperforms RoBERTa-Cont by 4.9/4.4, 5.5 and 1.7 absolute gains on three tasks, respectively.
Since RoBERTa-Sel-I is superior to RoBERTa-Sel-A on two tasks and three datasets, we further only present the results of RoBERTa-Sel-I combined with the Visibility Matrix mechanism.
The combination of selective mask policy Sel-I and visibility matrix (RoBERTa-Sel-I-VM) performs best in the LAMA Google RE, Closed-book QA and KG Reasoning.

Finally, we observe that at the same continual pre-training iterations, our models generally give higher accuracy than RoBERTa-Cont on all tasks, showing that our methods can also benefit in the efficiency of learning knowledge. In addition, though SSM introduces external tools (a trained NER tagger) and prior knowledge (expression to identify dates), our methods performs better than it. It is mainly because SSM only mask entities while leaves other kinds of tokens, which are also important for knowledge probing in the two task. SSM outperforms our methods on KG Reasoning, it is natural since KG Reasoning queries contain only entities and relations. 

\subsection{On Knowledge-Baring Tokens}
We also evaluate the continual pre-training on K-B tokens to see whether the improvement comes from the model's better understanding of K-B tokens. The evaluation data statistic is shown in Table~\ref{tab:analysis_data} (a) in Appendix.
The results are presented in Table~\ref{acc_on_kb}. From this table, we can see that our methods can help model better comprehend K-B tokens, showing that the overall better results in Table~\ref{results} comes models' comprehension of K-B tokens.

\subsection{Discovery on Invisible Tokens}

We find that the three tokens ``<s>'', ``</s>'' and ``.'' receiving much attention, consistently ranking on the top 20\%  in one piece of text. However, if we make  one or more of them invisible to other tokens, the performance on the three tasks will decrease by at least 5 points. Though they cannot be viewed as knowledge-baring tokens, they are still crucial for knowledge learning. We hypothesize they can store the general knowledge information of the text.


\section{Related Work}


\paragraph{Continual Pre-training of PLMs.}
\citet{gururangan-etal-2020-dont} reveals that continual pre-training on specific domains will contribute to the performance in downstream tasks within the same domains, and continual pre-training on some task's input data will also boost the performance on those datasets.   
\citet{REALM} proposed Salient span masking (SSM) which is using a NER tagger and rules to detect named entities and date, and then they mask at least one salient span each time when pretraining. On the contrary, we do not introduce any external information or prior knowledge to determine masks.
\citet{gu-etal-2020-train} first uses the training pairs of downstream tasks to help continue-pretrain a PLM. They find which tokens deleted from the input of task's training data will influence the confidence of prediction of the finetuned model, and they focus on masking those tokens when continual pre-training. 
\citet{ye-etal-2021-influence} proposed a two-loop meta-learned policy in continual pre-training BART for Closed-book QA Tasks, Knowledge-Intensive Tasks \citep{KILT} and abstractive summarization. They first continue to pre-train the BART with a passage and second train it with a (q,a) pair, and then they use the validation loss on the pair to update the parameters of mask policies.
The main difference between our work and the above two works is that their works use labelled datasets to help continual pre-training, while ours does not use any labelled data.


\paragraph{Knowledge Probing in PLMs.}
LAMA (LAnguage Model Analysis) \citep{LAMA} first uses the cloze-style test to evaluate how much knowledge in PLMs, it manually transfers some questions of SQuAD \citep{SQuAD} and some triples of Google RE, T-REx \citep{TREx} and ConceptNet \citep{ConceptNet} to cloze-style prompts. 
In this work, we create two variants x on LAMA probing and use the LAMA probing test and the variants to evaluate how much knowledge the model has learned from unstructured text. 

Despite increasing research in knowledge and PLMs, relatively less work associate knowledge from text with testing questions. \citet{how-much} and \citet{Fedus2021SwitchTS} use a set of query\&answer (QA) pairs to finetune the model and use another set of QA pairs to test it, which have no explicit correlation with pre-training data.  
We cannot exactly know whether the model learn from the training data or just solve questions by overlap between the finetuning data and test data \citep{overlap} or simply by spurious cues \citep{Probing}. In contrast, we  impose restrictions on the continual pre-training data and the test questions as well as get rid of finetuning process to ensure the model can only acquire needed knowledge from the training data.

\section{Conclusion}
We probe the behavior  of the pre-trained language models on unstructured text about the knowledge-baring and knowledge-free tokens, by asking those models to do the cloze-style test on our annotated data. We find that: (1) The model performs worse on K-B tokens; (2) The model gathers less attention on K-B tokens. To enable the model to better acquire knowledge from unstructured text, we consider two selective mask policies and adopt the visibility matrix mechanism to help the model focus on K-B tokens when learning from unstructured text. 
To our knowledge, we are the first to explore fully self-supervised learning of knowledge in continual pre-training.


\bibliography{anthology,custom}
\bibliographystyle{acl_natbib}

\clearpage

\appendix

\begin{table*}[t]
    \centering
    \scalebox{0.93}
    {\small
    \begin{tabular}{lc}
        \toprule
        \multicolumn{2}{c}{\textbf{{LAMA Probing}}} \\  \midrule
        \multicolumn{2}{p{16.6cm}}{\textbf{Train Text}: ... Kenya ranks \textcolor{red}{low} on Transparency International's Corruption Perception Index (CPI), a metric which attempts to gauge the prevalence of public sector corruption in various countries. ...} \\
        \textbf{Test Query}: On the CPI scale, Kenya ranks <mask>.  & \textbf{Test Answer}: \textcolor{red}{low} \\
        \midrule
        \multicolumn{2}{c}{\textbf{{Closed-book QA}}} \\  \midrule
        \multicolumn{2}{p{16.6cm}}{\textbf{Train Text}: ... The capital of the Ottoman empire was \textcolor{red}{Istanbul}....} \\
        \textbf{Test Query}: What was the capital of the Ottoman empire? <mask>  & \textbf{Test Answer}: \textcolor{red}{Istanbul} \\
        \midrule
        \multicolumn{2}{c}{\textbf{{KG Reasoning}}} \\
        \midrule
        \multicolumn{2}{p{16.6cm}}{\textbf{Train Text}: Shlomo Shriki, Israeli painter and artist, born in \textcolor{red}{Morocco} (1949), grew up and was educated in Kibbutz Yifat.} \\
        \textbf{Test Query}: Shlomo Shriki, place of birth, <mask>  & \textbf{Test Answer}: \textcolor{red}{Morocco} \\
        \bottomrule
    \end{tabular}
    }
    \caption{Examples of three tasks. The training text are all unstructured text and label-free. In validation/test, the model need to predict on the <mask> token.}
    \label{tasks_table}
    \vspace{-3mm}
\end{table*}

\begin{table}[t]
  \centering
  \small
  \setlength{\tabcolsep}{0.52mm}
  \begin{tabular}{c|c|c|c}
  \toprule
  & \makecell{Training\\Passages} 
  &\makecell{Validation\\Queries} 
  &\makecell{Testing\\Queries}\\
   \midrule
  \makecell{LAMA Probing\\(LAMA SQuAD)} & 271  &  152 &152\\
  \hline
  \makecell{LAMA Probing\\(LAMA Google RE)} & 5516
  & 2758 & 2758\\
   \midrule
  \makecell{Closed-book QA} & 271  &  152 &152\\
   \midrule
  \makecell{KG Reasoning} & 5516
  &  2206&2205\\
  \bottomrule
  \end{tabular}
  \caption{The statistics of three tasks (four datasets).
  }
  \vspace{-3mm}
  \label{task_data}
\end{table}

\begin{table}[t]
  \centering
  \small
  \begin{tabular}{c|c|c}
  \toprule
  & \makecell{\#Sentences} &\makecell{\#Masked Tokens} \\
  \midrule
  \makecell{LAMA SQuAD\\(Knowledge-Baring)} & 609 & 739\\
  \makecell{LAMA SQuAD\\(Knowledge-Free)} & 524 & 532\\
  \midrule
  \makecell{LAMA Google RE\\(Knowledge -baring)} & 1268 & 1715\\
  \midrule
  \makecell{LAMA Google RE\\(Knowledge -free)} & 865 & 975\\
  \midrule
  \multicolumn{3}{c}{\makecell{(a) Data after every token of the 200 samples\\ is masked separately, which is used for\\ accuracy analysis in Section~\ref{sec:accuracy_analysis}.}}\\
  
  \midrule
  & \makecell{\#Sentences} &\makecell{\# Tokens} \\
  \midrule
  \makecell{LAMA SQuAD} & 100 & 1471\\
  \midrule
  \makecell{LAMA Google RE} & 100
  & 2903\\
  \midrule
  \multicolumn{3}{c}{\makecell{(b) Data used for attention analysis\\ in Section~\ref{sec:attention_analysis}.}} \\
  \bottomrule
  \end{tabular}
  \caption{Data statistics after the 200 samples are processed for analysis in Section~\ref{sec:accuracy_analysis} and Section~\ref{sec:attention_analysis}.
  }
  \vspace{-2mm}
  \label{tab:analysis_data}
\end{table}

\begin{table}[t]
  \centering
  \small
  \setlength{\tabcolsep}{0.52mm}
  \begin{tabular}{l|c}
  \toprule
  \textbf{Hyperparam} &  \\\hline
  Learning Rate& 1e-4 \\
  Train Batch Size & 256 (passages)\\
  MLM propability & 0.15\\
  Max Tokens Length & 512\\
  Optimizer & Adam\\
  Adam $\epsilon$ &1e-6\\
  Adam $\beta_{1}$ & 0.9\\
  Adam $\beta_{2}$ & 0.98\\
  Weight Decay &0.01\\
  Learning Rate Decay &Linear\\
  \bottomrule
  \end{tabular}
  \caption{The hyper-parameters for continual pre-training RoBERTa in this work.
  }
  \label{hyper_para}
  \vspace{-2mm}
\end{table}

\begin{table*}[t]
  \centering
  \small
  \setlength{\tabcolsep}{0.6mm}
  \begin{tabular}{c|c|c|c|c|c|c|c|c|c|c|c|c|c|c|c}
  \toprule
\makecell{Knowledge-\\Baring\\Tokens}
&\makecell{NN\\243\\0.329}
&\makecell{NNP\\179\\0.242}
&\makecell{JJ\\68\\0.092}
&\makecell{NNS\\49\\0.066}
&\makecell{VBN\\46\\0.062}
&\makecell{VBD\\21\\0.028}
&\makecell{CD\\20\\0.027}
&\makecell{VBZ\\19\\0.026}
&\makecell{VB\\15\\0.02}
&\makecell{IN\\14\\0.019}
&\makecell{POS\\8\\0.011}
&\makecell{RB\\7\\0.009}
&\makecell{NNPS\\6\\0.008}
&\makecell{VBP\\5\\0.007}
&\makecell{VBG\\5\\0.007}\\
  \midrule
 \makecell{Knowledge-\\Free\\Tokens}
&\makecell{IN\\149\\0.28}
&\makecell{DT\\109\\0.205}
&\makecell{.\\100\\0.188}
&\makecell{VBZ\\39\\0.073}
&\makecell{VBD\\31\\0.058}
&\makecell{TO\\14\\0.026}
&\makecell{,\\13\\0.024}
&\makecell{CC\\9\\0.017}
&\makecell{VB\\8\\0.015}
&\makecell{RB\\8\\0.015}
&\makecell{WDT\\7\\0.013}
&\makecell{PRP\$\\6\\0.011}
&\makecell{VBP\\5\\0.009}
&\makecell{WRB\\4\\0.008}
&\makecell{MD\\4\\0.008}\\
  \midrule
  \multicolumn{16}{c}{(a) In LAMA SQuAD samples} \\
  \midrule
  \makecell{Knowledge-\\Baring\\Tokens}
&\makecell{NNP\\657\\0.383}
&\makecell{NN\\419\\0.244}
&\makecell{CD\\205\\0.12}
&\makecell{VBN\\102\\0.059}
&\makecell{JJ\\91\\0.053}
&\makecell{IN\\38\\0.022}
&\makecell{VBD\\36\\0.021}
&\makecell{NNS\\23\\0.013}
&\makecell{FW\\22\\0.013}
&\makecell{PRP\\19\\0.011}
&\makecell{VBG\\14\\0.008}
&\makecell{DT\\12\\0.007}
&\makecell{VBZ\\11\\0.006}
&\makecell{VBP\\10\\0.006}
&\makecell{RB\\9\\0.005}
  \\\midrule
\makecell{Knowledge-\\Free\\Tokens}
&\makecell{IN\\230\\0.236}
&\makecell{,\\143\\0.147}
&\makecell{DT\\113\\0.116}
&\makecell{.\\110\\0.113}
&\makecell{CC\\58\\0.059}
&\makecell{-RRB-\\58\\0.059}
&\makecell{-LRB-\\58\\0.059}
&\makecell{VBD\\53\\0.054}
&\makecell{VBZ\\40\\0.041}
&\makecell{HYPH\\26\\0.027}
&\makecell{:\\18\\0.018}
&\makecell{RB\\16\\0.016}
&\makecell{WP\\11\\0.011}
&\makecell{PRP\$\\11\\0.011}
&\makecell{WRB\\4\\0.004}
  \\\midrule
  \multicolumn{16}{c}{(b) In LAMA Google RE samples}\\
  \bottomrule
  \end{tabular}
  \caption{ Part-of-speech Results on our annotated samples. For each cell, the tag name is at the top, the number of this tag is in the middle, the proportion of this tag is in the bottom. For each type of token in each data set, we only display the top-15 tags.
  }
  \label{pos_on_samples}
  \vspace{-2mm}
\end{table*}

\begin{table*}[t!]
  \centering
  \small
  \setlength{\tabcolsep}{0.6mm}
  \begin{tabular}{c|c|c|c|c|c|c|c|c|c|c|c|c|c|c|c}
  \toprule
\makecell{RoBERTa-Cont\\(Random)}
&\makecell{NNP\\0.206}
&\makecell{NN\\0.151}
&\makecell{IN\\0.117}
&\makecell{DT\\0.067}
&\makecell{CD\\0.054}
&\makecell{JJ\\0.048}
&\makecell{,\\0.047}
&\makecell{VBN\\0.045}
&\makecell{VBD\\0.043}
&\makecell{.\\0.042}
&\makecell{VBZ\\0.031}
&\makecell{NNS\\0.021}
&\makecell{CC\\0.019}
&\makecell{-RRB-\\0.015}
&\makecell{RB\\0.013}\\
  \midrule
 \makecell{RoBERTa-Sel-I}
&\makecell{NNP\\0.24}
&\makecell{NN\\0.181}
&\makecell{IN\\0.097}
&\makecell{CD\\0.073}
&\makecell{DT\\0.051}
&\makecell{JJ\\0.047}
&\makecell{VBN\\0.038}
&\makecell{VBD\\0.037}
&\makecell{,\\0.034}
&\makecell{.\\0.033}
&\makecell{VBZ\\0.024}
&\makecell{NNS\\0.023}
&\makecell{CC\\0.016}
&\makecell{-RRB-\\0.012}
&\makecell{RB\\0.012}\\
  \midrule
  \makecell{RoBERTa-Sel-A}
&\makecell{NNP\\0.265}
&\makecell{NN\\0.167}
&\makecell{IN\\0.112}
&\makecell{CD\\0.086}
&\makecell{JJ\\0.049}
&\makecell{DT\\0.048}
&\makecell{VBN\\0.033}
&\makecell{,\\0.026}
&\makecell{VBD\\0.024}
&\makecell{.\\0.021}
&\makecell{VBZ\\0.021}
&\makecell{NNS\\0.018}
&\makecell{CC\\0.013}
&\makecell{-LRB-\\0.012}
&\makecell{RB\\0.012}\\
  \bottomrule
  \end{tabular}
  \caption{The result of part-of-speech analysis for three mask policies. For each cell, the tag name is at the top and the proportion of this tag is in the bottom.
  }
  \label{pos_on_masks}
  \vspace{-2mm}
\end{table*}

\begin{table}[t]
  \centering
  \small
  \setlength{\tabcolsep}{0.52mm}
  \begin{tabular}{l|c}
  \toprule
  \textbf{Method} & K-B / K-F ratio \\
  \midrule
  RoBERTa-Cont (Random)& 1.47:1 \\
  RoBERTa-Sel-I & 2.16:1\\
  RoBERTa-Sel-A & 2.33:1\\
  \bottomrule
  \end{tabular}
  \caption{The K-B to K-F ratios for three mask policies. The experiment is conducted on the samples which are annotated with K-B and K-F.
  }
  \label{kbf_on_masks}
  \vspace{-2mm}
\end{table}

\section{Hyper-parameters}
\label{sec:hyper_para}
The traditional hyper-parameters for continual pre-training RoBERTa can be seen at Table~\ref{hyper_para}.

Moreover, for RoBERTa-Sel-I and RoBERTa-Sel-A, we set the first-phase MLM probability as 15\% and the second-phase MLM probability as 10\%.
For the RoBERTa-SSM, we adopt a publicly released NER model, which is based on RoBERTa-base and trained on the conll2003 dataset,~\footnote{huggingface.co/andi611/roberta-base-ner-conll2003} and a regular expression to identify named entities and date, respectively.
In the LAMA Probing task, all models are trained for 100 iterations. For the visible mechanism, we use the original RoBERTa-large to find the knowledge-free tokens.
In the Closed-book QA task, models are trained for 500 iterations.
For the visible matrix mechanism, we set $\tau$ as 3 for the two datasets.

\section{Details of POS analysis on Samples}

We present a detailed results of part-of-speech tagging analysis of annotated samples in Table~\ref{pos_on_samples}.

\section{Mask Analysis}

To compare three different mask policies, namely RoBERTa-Cont,  RoBERTa-Sel-I and RoBERTa-Sel-A
we conduct 10-iteration continual pre-training on the 200 samples in Section~\ref{sec:knowledge_section} and record their masked tokens.

Then We take part-of-speech analysis on the mask tokens for the three mask policies, which is presented in Table~\ref{pos_on_masks}. From the result, we can see that our two selective mask policies choose more nouns, numbers, verbs and adjective words to mask than the random mask policies.

We also calculate the K-B / K-F ratio of masked tokens for the three mask policies and  list the result in Table~\ref{kbf_on_masks}.
From the table, it can be seen that our two selective mask policies can significantly increase the proportion of K-B tokens in the masked tokens.

\end{document}